\documentclass[conference]{IEEEtran}
\IEEEoverridecommandlockouts
\usepackage{stfloats}
\usepackage[T1]{fontenc}
\usepackage{amsmath,amssymb,amsfonts}
\usepackage{mathrsfs}  
\usepackage{algorithmic}
\usepackage{subfig}
\usepackage{graphicx}
\usepackage{textcomp}
\usepackage{balance}

\def\BibTeX{{\rm B\kern-.05em{\sc i\kern-.025em b}\kern-.08em
    T\kern-.1667em\lower.7ex\hbox{E}\kern-.125emX}}

\usepackage{multirow}

\graphicspath{{./figures}}

\begin{document}

\title{Risk-Aware Reward Shaping of Reinforcement Learning Agents for Autonomous Driving}

\author{Lin-Chi Wu$^{1}$, Zengjie Zhang$^{1*}$, Sofie Haesaert$^{1}$, Zhiqiang Ma$^{2*}$, and Zhiyong Sun$^{1}$%
\thanks{This work was supported by the European project SymAware under grant No. 101070802.} 
\thanks{$^{1}$Lin-Chi Wu, Zengjie Zhang, Sofie Haesaert, and Zhiyong Sun are with the Department of Electrical Engineering, Eindhoven University of Technology, Netherlands. {\tt\footnotesize l.wu@student.tue.nl, \{z.zhang3, s.haesaert, z.sun\}@tue.nl}.} 
\thanks{$^{2}$Zhiqiang Ma is with Research Center for Intelligent Robotics, School of Astronautics, Northwestern Polytechnical University, Xi'an 710072, China, {\tt\footnotesize zhiqiangma@nwpu.edu.cn}.}%
\thanks{* Common corresponding authors.}
}

\maketitle

\begin{abstract}
Reinforcement learning (RL) is an effective approach to motion planning in autonomous driving, where an optimal driving policy can be automatically learned using the interaction data with the environment. Nevertheless, the reward function for an RL agent, which is significant to its performance, is challenging to determine. The conventional work mainly focuses on rewarding safe driving states but does not incorporate the awareness of risky driving behaviors of the vehicles. In this paper, we investigate how to use risk-aware reward shaping to leverage the training and test performance of RL agents in autonomous driving. Based on the essential requirements that prescribe the safety specifications for general autonomous driving in practice, we propose additional reshaped reward terms that encourage exploration and penalize risky driving behaviors. A simulation study in OpenAI Gym indicates the advantage of risk-aware reward shaping for various RL agents. Also, we point out that proximal policy optimization (PPO) is likely to be the best RL method that works with risk-aware reward shaping. 
\end{abstract}


\section{Introduction}
Autonomous driving is dedicated to developing an automatic driving system that can substitute human drivers with equivalent performance. An autonomous driving policy should not only avoid severe events like collisions or rule violations but also be aware of upcoming risky events and take action in advance. 
A typical human driving pattern has been abstracted into a decision-making system that contains four layers: route planning, behavioral layer, motion planning, and local feedback control~\cite{aradi2020survey}. The decision maker in each layer is designed to accomplish the driving task subject to certain rules defined at different levels. For example, a route planner is restricted by urban terrain and road conditions. In the behavioral layer, the vehicles should follow certain traffic rules. Also, motion planning of a vehicle is concerned with collision avoidance with obstacles and other vehicles. In industrial applications, however, autonomous driving is categorized into six levels of behaviors~\cite{paden2016survey, sae2018taxonomy}. According to the National Highway Traffic Safety Administration in the United States, the autonomous system product that is permitted on the road is of level 2 \cite{national_highway_traffic_safety_administration_evolution_2022}, which means that the system still requires additional assistance from a human expert who perceives the traffic condition behind the wheel. 

Reinforcement learning (RL) provides an effective way to automatically learn a driving policy from the interaction with the environment, such as deep-Q-network (DQN)~\cite{chae2017autonomous}, deep deterministic policy gradient (DDPG)~\cite{kiran2021deep}, and proximal policy optimization (PPO)~\cite{wu2021deep}. RL does not require human demonstrations but the data obtained from the interaction with the environment during the training process~\cite{kiran2021deep, osinski2020simulation}. The core idea of learning to drive is about trial and error like the agent playing a game with the environment. Corresponding to the multi-layer design of an autonomous driving system mentioned above, hierarchical RL is popularly applied by decomposing the entire driving task into several subtasks~\cite{hsu2023deep, gonzalez2015review}. The decomposition of the driving task is heuristic and usually requires engineering experience and knowledge from experts. Proper reward shaping allows the agent to perform better in complex scenarios \cite{gutierrez2022reinforcement, ye2020automated}. Besides, heuristic models, such as motion primitives, are also used to simplify the learning process~\cite{pedrosa2022learning, sheckells2017fast}. Different from heuristic RL, end-to-end RL uses a deep neural network to learn an overall driving policy that directly maps vision perception to control commands~\cite{xiao2020multimodal, amini2020learning}. A review of RL methods used to solve autonomous driving problems can be found in~\cite{kiran2021deep}. However, ensuring a risk-free system, i.e., one that does not make harmful or unethical decisions, is still challenging due to the difficulty of describing risky driving behaviors. As a result, one may find that RL-based vehicles often generate aggressive and reckless driving behaviors since the reward does not fully incorporate the potential risks of these behaviors.

\begin{figure}[htbp]
   \centering
   \subfloat[close to edge]{\includegraphics[width=0.16\textwidth]{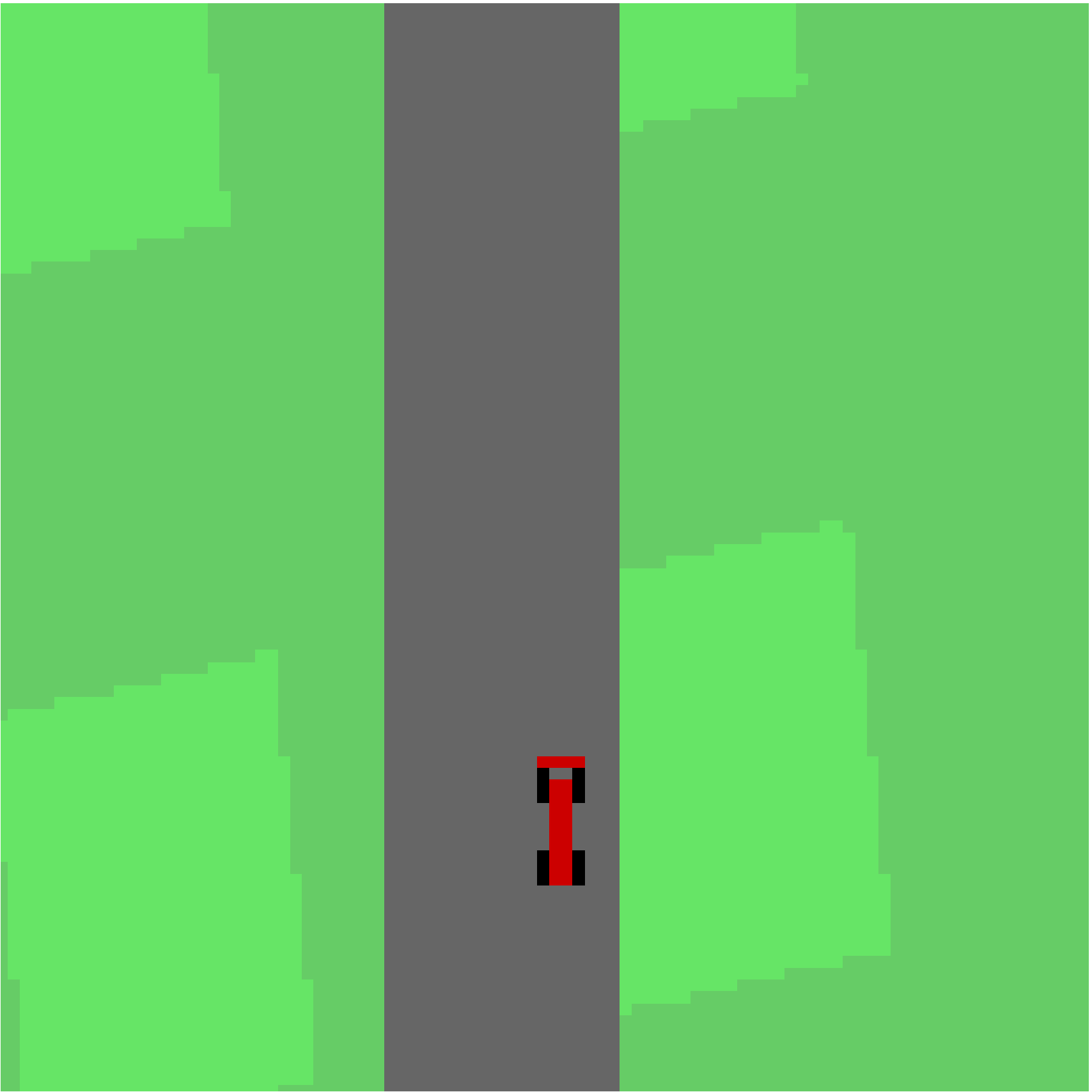}\label{fig:risk_1}}
   \hfill
   \subfloat[obstacle in front]{\includegraphics[width=0.16\textwidth]{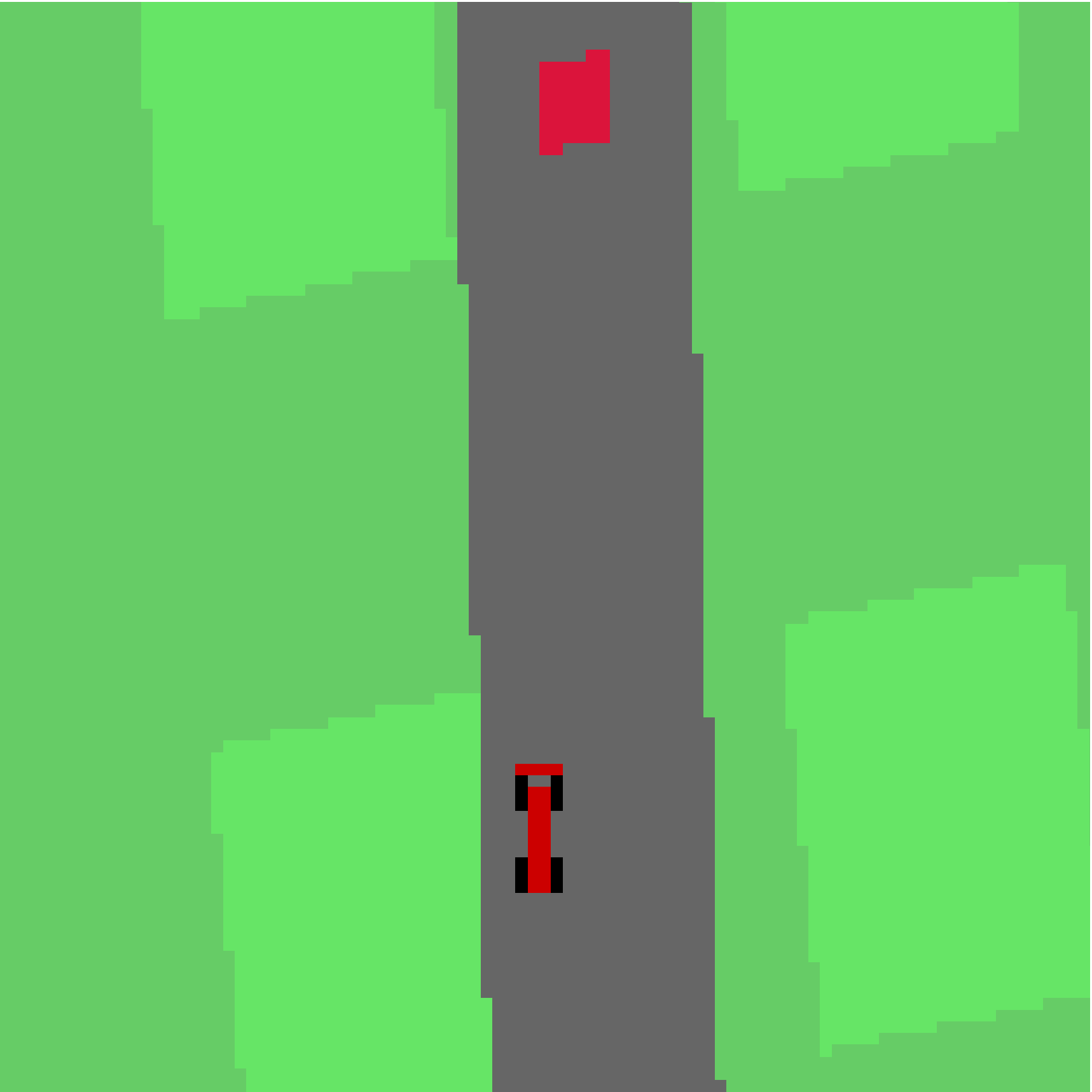}\label{fig:risk_2}}
   \hfill
   \subfloat[close to the turn]{\includegraphics[width=0.16\textwidth]{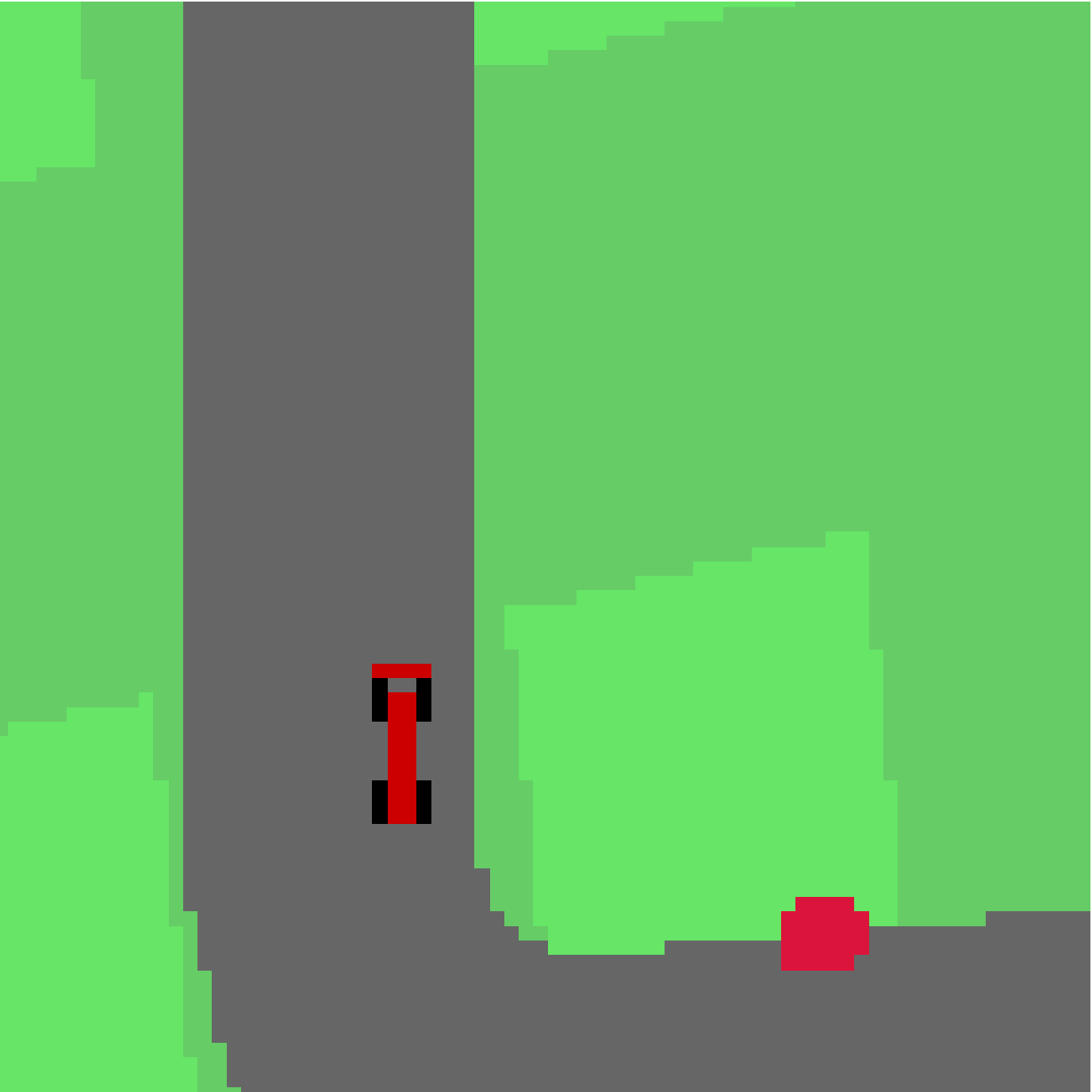}\label{fig:risk_3}}
   \caption{The illustration of risky events.}
   \label{fig:risk}
\end{figure}

Risks usually refer to events that are not fatal to the system for the current moment but may lead to severe consequences in the future. Fig.~\ref{fig:risk} illustrates several examples of risky events obtained from the \textit{CarRacing-V0} environment provided by OpenAI Gym. Fig.~\ref{fig:risk_1} shows the case where the car is driving close to the edge of the road, which makes it more likely that the car runs out of the road, compared to driving in the center of the road. Fig.~\ref{fig:risk_2} shows that an obstacle is in front of the car but it still takes a while until the collision actually happens. Fig.~\ref{fig:risk_3} indicates that the car is close to the turn where the car is under the threat of vision occlusion. Different from fatal events like collisions, autonomous vehicles under risk are still in safe conditions, but they may run into fatal events in the short future. Vehicles under risk still have the chance to take conservative actions to prevent possible fatal events. Thus, how to be aware of risks and taking proper actions to prevent fatal events is an effective way of avoiding severe accidents. Risk-aware technologies have been widely applied to robotic systems, such as estimating external disturbances~\cite{zhang2019integral, zhang2022disturbance}, predicting the likelihood of collisions~\cite{zhang2020online}, and handling instantaneous perception~\cite{li2021instantaneous}. It is very recently that risk-awareness has become an important topic for autonomous driving~\cite{huang2018hybrid}. Existing work has integrated the risk-awareness procedure into model predictive control (MPC) to solve the motion planning problem for autonomous driving~\cite{kim2022physics}. In~\cite{nyberg2021risk}, risk awareness has been encoded as rule-based specifications. In~\cite{khonji2020risk}, a stochastic uncertainty model is constructed to quantify risks. However, risk awareness has not been introduced to reinforcement learning (RL) for the training of autonomous driving policies. This can be achieved by encoding the risk-aware specifications into the rewards of the RL agent. Nevertheless, this is a heuristic process where empirical knowledge of reinforcement learning is needed.

In this paper, we investigate how to use risk-aware reward shaping to improve the capability of RL-trained autonomous vehicles to avoid risky driving behaviors while ensuring sufficient exploration of the environment. Additional reward terms are introduced to encourage explorative driving directions and penalize risky driving behaviors. The reshaped reward is applied to three different agent models, namely a DQN agent, a DDPG agent, and a PPO agent, to validate its general applicability. A comparison study in the OpenAI Gym environment validates that the agents trained with the reshaped reward show better training performance and higher test scores than the ones with the non-reshaped reward. The different effects of the reshaped reward terms on different agents are also addressed, which supports that PPO is the most suitable method for the reshaped reward in the studied case. In general, the main contributions of this paper are summarized as follows.
\begin{itemize}
\item Encoding risk-aware specifications into RL for autonomous driving via reward shaping.
\item Comparison studies to show to which RL approach the reshaped reward is most effective.
\end{itemize}

The rest of the paper is organized as follows.
Sec.~\ref{sec:pre} provides the preliminary knowledge of this paper. Sec.~\ref{sec:main} presents our proposed reward-shaping method. Sec.~\ref{sec:exp} conducts a simulation study to validate the efficacy of the reshaped risk-aware reward. Sec.~\ref{sec:con} concludes the paper.

\section{Preliminaries and Problem Statement}\label{sec:pre}

This section introduces the preliminary knowledge of this paper, including the formulation of reinforcement learning and the OpenAI Gym simulation environment. Then, we formulate the main problem to be studied in the paper.

\subsection{Reinforcement Learning (RL)}
Motion planning in autonomous driving can be solved using RL through the formulation of a Markov Decision Process (MDP). An MDP is a tuple $M=(S,A,\mathscr{F}, \mathscr{R})$, where $S$ and $A$ are the spaces of states and actions of the agent, $\mathscr{R}:S \times A \rightarrow \mathbb{R}$ is the reward function, and $\mathscr{F}(s_t'|s_t, a_t)$ is a distribution that denotes the state transition of the environment, where $s_t, s_t' \in S$ and $a_t \in A$ are, respectively, the current state, the successive state, and the action of the environment at certain time $t$, where $t=0,1,\cdots, T$ and $T \in \mathbb{R}^+$ is the length of the trajectory. The objective of RL is to solve an optimal policy $\pi(s_t|a_t)$ such that the following accumulated reward $\displaystyle J(\pi) = \mathop{\mathsf{E}}_{a \sim \pi}\!\left(\sum_{t=0}^T\! \gamma^{t} \mathscr{R}(s_t, a_t)\right)$ is maximized, where $0 < \gamma <1$ is a discount factor. The optimal policy can be solved with either off-policy methods, such as DQN and DDPG, or on-policy methods like PPO. DQN and DDPG use deep neural networks to approximate the Q-value function $\displaystyle Q^*(s, a) = \max_{\pi} \mathop{\mathsf{E}}_{a\sim\pi}\!\left( \mathscr{R}(s, a) \right)$, $s \in S$, 
which represents the expected future reward for the current policy $\pi$ at a certain state $s \in S$. 
Both approaches approximate the Q-value function using the Bellman equation, $\displaystyle Q^*(s,a)=\mathop{\mathsf{E}}\limits_{s' \sim \mathscr{F}} \!\left[( \mathscr{R}(s,a)+\gamma \max_{a'} Q^*(s',a') \right]$ which renders an iterative update law. The details of DQN and DDPG methods can be referred to in~\cite{tan2021reinforcement}. Different from them, PPO solves the optimal policy using a policy-gradient search method~\cite{zhang2023research}.

\subsection{Problem Statement}\label{sec:pb}

\begin{figure}[htbp]
    \centering
    \includegraphics[width=0.8\linewidth]{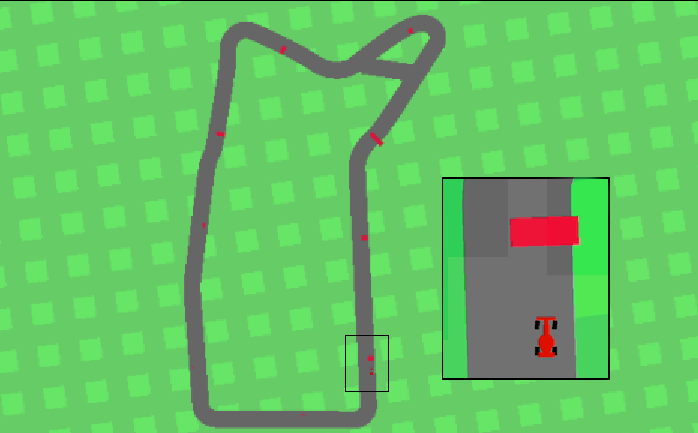}
    \caption{A screenshot of the \textit{CarRacing-V0} environment, where the track is marked as gray, the green area is the grassland around the track, the red blocks denote the obstacles, and the racing car is marked with a red body with black wheels.}
    \label{fig:map}
\end{figure}

In this paper, we consider an autonomous driving task performed in the \textit{CarRacing-V0} environment provided by the OpenAI Gym box2D environment sets~\cite{brockman_openai_2016}. The environment renders a racing car that can run through a 2D map, as shown in Fig \ref{fig:map}. More details about this map can be referred to in~\cite{noauthor_improvement_nodate}.
In this environment, the car is required to drive along a track with the following requirements satisfied.
\medskip \\
\textbf{Autonomous driving requirements:} 
\begin{itemize}
\item The vehicle should drive along the track and stay inside the track. It fails to survive when driving out of the track.
\item The vehicle should avoid collisions with any obstacles showing up on the track. It fails when hitting an obstacle.
\end{itemize}

An MDP $M=(S,A,\mathscr{F}, \mathscr{R})$ is defined to train an RL agent to achieve these requirements. The state space $S$ is the set of all images showing a bird's view of a car at the bottom middle of the picture with a pixel dimension of $(96, 96)$ in RGB format, saved in the format of grayscale. The action set $A$ contains five discrete actions, namely \textit{no action}, \textit{steer left}, \textit{steer right}, \textit{accelerate}, and \textit{brake}. The state transition $\mathscr{F}$ is prescribed by the OpenAI Gym environment. The main problem to be solved in this paper is to define the reward function $\mathscr{R}$, such that the solved policy $\pi$ achieves the above-mentioned autonomous driving requirements.

\section{Reward Shaping for Autonomous Driving}\label{sec:main}

This section presents our main methods of risk-aware reward shaping of RL agents. We first extract the essential requirements for autonomous driving from practical experience. Then, we introduce the additional reshaped reward terms based on the encouragement of explorative driving intentions and the penalization of risky driving behaviors.

\subsection{Essential Principles of Reward Shaping}

As addressed in Sec.~\ref{sec:pb}, the essential objective of the problem is to train an RL agent that drives along the track without collisions with the obstacles on the track. The encoding of safety requirements, such as collision avoidance, has been well-studied in the existing RL methods. However, risky driving behaviors are rarely incorporated by the conventional reward-shaping approaches. For autonomous driving, a risk may not be fatal but should not last for a long time. An example of a risky event might be when the car drives too close to the edges of the track. In this case, the car is still safe but it has a higher possibility of running out of the track than driving in the center of the track. Thus, a proper action of the vehicle is to drive back to the center of the track instead of maintaining the original direction. In the meantime, RL requires that the agent performs a reasonable balance between the exploration of the environment and the exploitation of the historical interaction with the environment. Getting stuck in the visited states is also a risk that prevents the agent from sufficiently exploring the environment. Incorporating driving risks into motion planning is promising to effectively improve the reliability of driving. Specifically, positive rewards should be added for explorative driving intentions that lead to unvisited states. On the contrary, risky driving behaviors should receive mild negative rewards to encourage risk mitigation. These principles are explained in detail as follows. 

\subsection{Encouragement of Explorative Driving Intensions}\label{sec:enc}

The training of the RL agent should encourage the exploration of the car in the environment. Specifically, the car should try to sufficiently explore the environment by driving along new paths. This is a practical principle to avoid the agent being stuck in the visited states. The OpenAI Gym provides an internal API to check whether a state has been visited in the history. We add a positive reward $r^{\mathrm{exp}}_t>0$ whenever the current state is recognized as \textit{unvisited}. The principle is summarized as follows.
\medskip \\
\noindent \textbf{Encouragement principle:}
\begin{itemize}
\item The agent receives a positive reward $r^{\mathrm{exp}}_t$ when the agent is visiting the new path.
\end{itemize}

\subsection{Penalty of Risky Driving}\label{sec:penal}

Alongside encouragement, penalties are also needed to discourage dangerous or risky driving behaviors. In the literature, penalties are mostly raised for collision avoidance since it is the biggest concern of autonomous driving. Similar to most of the existing work, we impose a considerable negative reward $r^{\mathrm{obs}}_t <0$ on the agent when it collides with an obstacle on the track. Moreover, we also consider the case where the car goes out of the track and drive into the grassland. This case is a risk but not a fatal event since the agent can still survive and come back to the track within a finite time. Therefore, different from the collisions with the obstacles, we penalize the agent for going out with a mild negative reward $r^{\mathrm{risk}}_t < 0$. We also imposed a soft penalty of $r^{\mathrm{alive}}_t <0$ for each step where the agent remains inside the track, hoping to stimulate the agent to earn a positive reward by exploring new paths. The penalty principles are summarized as follows.
\medskip \\
\noindent \textbf{Penalization principle:}
\begin{itemize}
\item A considerable penalty $r^{\mathrm{obs}}_t <0$ is given to the agent when the car collides with the obstacle.
\item A mild penalty $r^{\mathrm{risk}}_t < 0$ is given when the agent encounters risky events, i.e., getting close to the edges.
\item A mild penalty $r^{\mathrm{alive}}_t<0$ is added for each episode step.
\end{itemize}

\subsection{Liveness Reshaping}

Based on the encouragement and the penalization principles introduced in Sec.~\ref{sec:enc} and Sec.~\ref{sec:penal}, we propose the following risk-aware reward function for the training of an autonomous driving agent,
\begin{equation}\label{eq:reward_config}    
\begin{split}
    \mathscr{R}(s_t)=\,& r^{\mathrm{exp}}_t + r^{\mathrm{obs}}_t + r^{\mathrm{risk}}_t + r^{\mathrm{alive}}_t,
\end{split}
\end{equation}
where the reward terms $r^{\mathrm{exp}}_t$, $r^{\mathrm{obs}}_t$, $r^{\mathrm{risk}}_t$, $r^{\mathrm{alive}}_t$ are only added to \eqref{eq:reward_config} when their corresponding principles in Sec.~\ref{sec:enc} and Sec.~\ref{sec:penal} are true.

\begin{figure}[htbp]
    \centering
\includegraphics[width=0.8\linewidth]{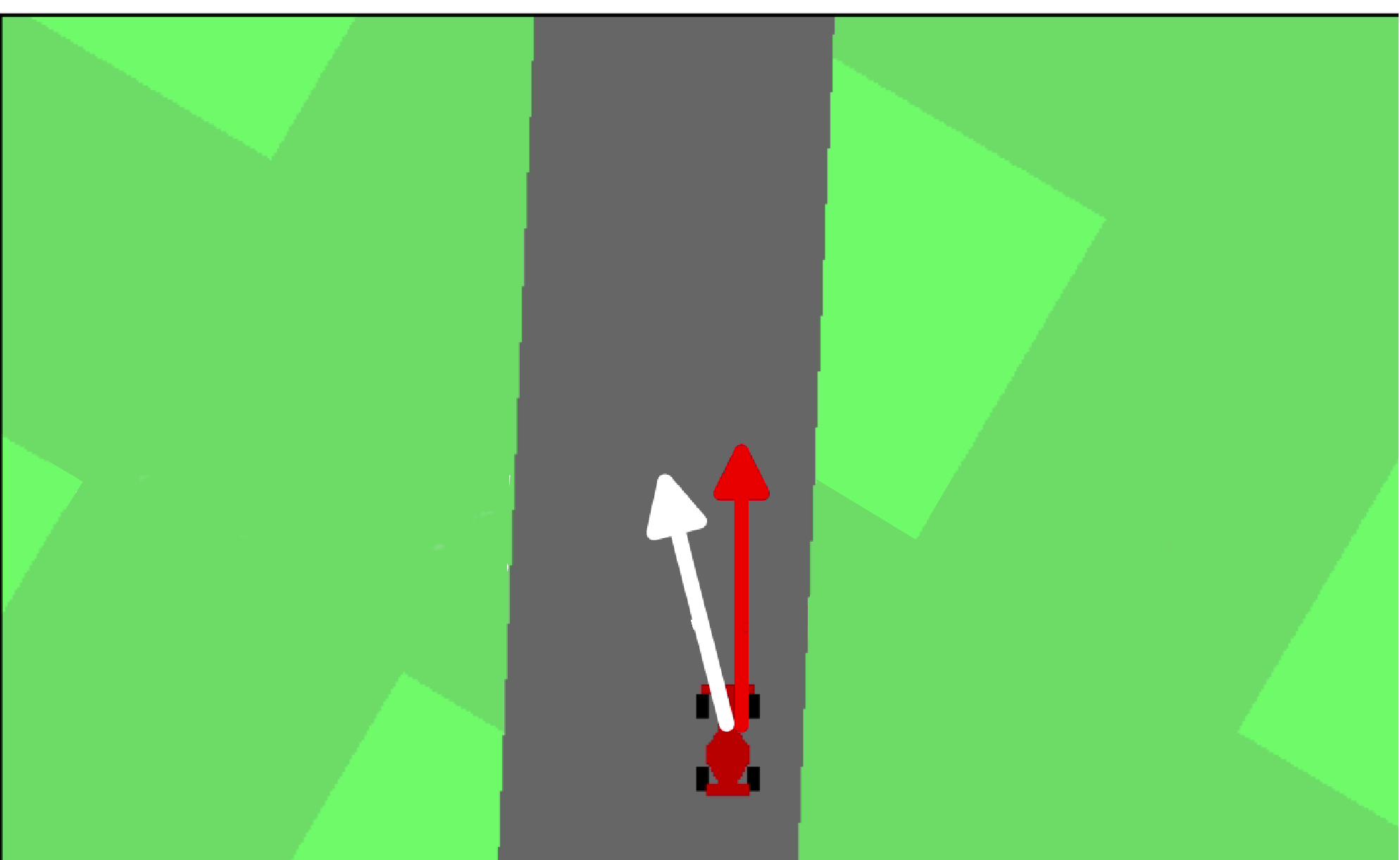}
    \caption{Short liveness may produce an agent that drives along the diagonal direction of the track (white arrow) instead of moving straightforwardly along the track (red arrow).}
    \label{fig:diagonal_case}
\end{figure}

The liveness of the agent, i.e., the number of steps of each episode should also be carefully determined. If the liveness of the agent is too short, the agent may be \textit{short-sighted} as to lose the capability to evaluate risks. For example, an agent with short liveness tends to drive along the diagonal direction of the track (white arrow in Fig.~\ref{fig:diagonal_case}) instead of the parallel direction along the track bank (red arrow in Fig.~\ref{fig:diagonal_case}) since it can not distinguish between the possible consequences of the two directions. In other words, it can not foresee the risk of going out of track after a longer time duration. In this sense, the length of liveness $N_{\mathrm{eps}}$ should be properly selected. Moreover, the duration of risks should also be limited. Therefore, we propose the following termination conditions to prescribe the liveness of the car.
\medskip \\
\noindent \textbf{Termination conditions:}
\begin{itemize}
\item The car collides with an obstacle.
\item The episode reaches the number limit $N_{\mathrm{eps}}$ of steps.
\item The risky event lasts for more than a fixed time $T_{\mathrm{risk}}$.
\item The accumulated reward reaches a positive value $R_{\mathrm{up}}$.
\item The accumulated reward reaches a negative value $R_{\mathrm{down}}$.
\end{itemize}

\section{Experimental Studies}\label{sec:exp}

In this section, the performance of the agents with the reward shaping is evaluated in the \textit{CarRacing-V0} simulation experiment. We will also discuss the strengths and weaknesses of different RL agents with the reshaped reward function. The code of the experimental studies is available at~\cite{Wu_Risk-Aware_Reward_Shaping_2023}.

\subsection{Experiment Configuration}

\begin{figure}[htbp]
    \centering
    \includegraphics[width=0.9\linewidth]{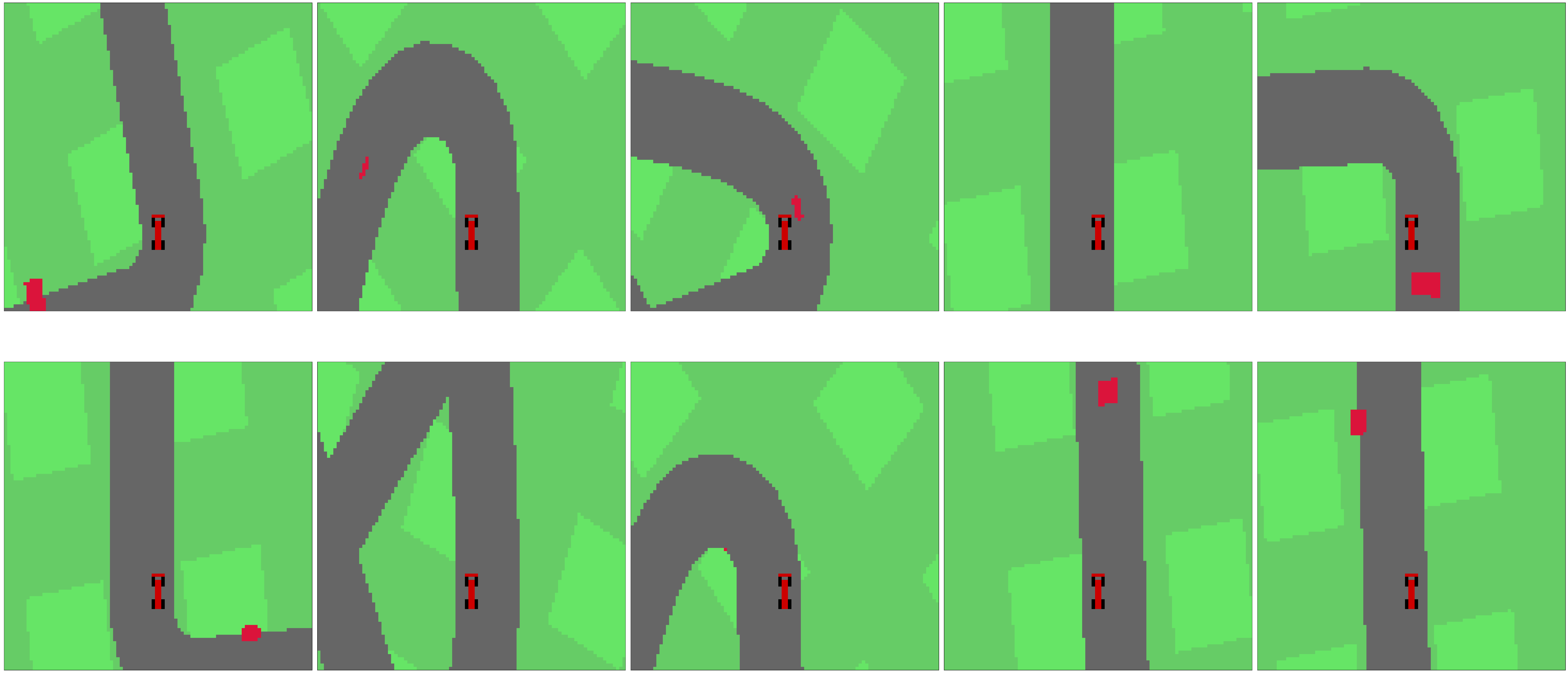}
    \caption{The random initial conditions of training and test.}
    \label{fig:Random_place}
\end{figure}

The experiment study is run on a Micro-Star International (MSI) laptop with a 12th Intel i7-12700H CPU and an NVIDIA GeForce RTX 3070 Laptop GPU. The initial positions of the car are randomly sampled. Ten possible initial conditions are shown in Fig.~\ref{fig:Random_place} as examples. The randomness of the initial locations can help the agent to explore the environment sufficiently. The sampling rate of the environment is $50$ frames per second, and the action is performed at each frame sampling. Therefore, the average sampling time of the environment is $0.02\,$s.


To address the importance of each reward term in~\eqref{eq:reward_config}, we make a comparison between the default and the reshaped reward functions. The values of the reward terms for the two reward functions are presented in Table~\ref{tab:reward_config}. Compared to the default reward function, the reshaped reward function enlarges the absolute values of $r^{\mathrm{exp}}_t$ and $r^{\mathrm{risk}}_t$, aiming at valuing more on the encouragement of exploration and the penalization on driving risks. Besides, the reshaped-reward agent is given a larger episode size to make it longer-sighted. The collision-avoidance and the liveness reward $r^{\mathrm{obs}}_t$ and $r^{\mathrm{alive}}_t$ are set the same for both reward functions since their functionalities have been well studied in the literature. Also, for both cases, the risk tolerance time is $T_{\mathrm{risk}} = 5\,$s, and the reward limits are $R_{\mathrm{up}} = 3000$ and $R_{\mathrm{down}}=-400$. The comparison study is performed on three different RL agents, namely a DQN agent, a DDPG agent, and a PPO agent.



\linespread{1.2}
\begin{table}[htbp]
\centering
\caption{The default and the reshaped reward functions}
\label{tab:reward_config}
\begin{tabular}{c|c|c|c|c|c}
\hline
reward function& $r^{\mathrm{exp}}_t$ & $r^{\mathrm{obs}}_t$ & $r^{\mathrm{risk}}_t$ & $r^{\mathrm{alive}}_t$ & $N_{\mathrm{eps}}$ \\
\hline
default & +1 & -600 & -1 & -1 & 700 \\
reshaped & +1.4 & -600 & -200 & -1 & 1200 \\
\hline
\end{tabular}
\end{table}
\linespread{1}

\subsection{Training Results}\label{sec:train_result}

\begin{figure}[htbp] 
\centering
\subfloat[]{
\label{fig:rewar1}\includegraphics[width=0.46\linewidth]{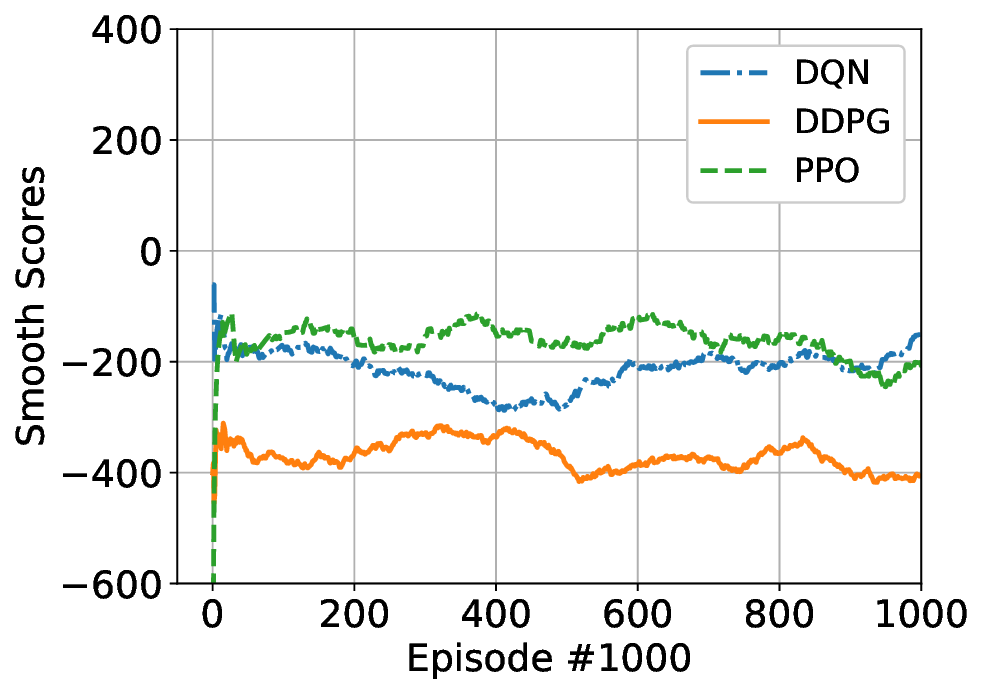}
    }
\hfill
\subfloat[]{
\label{fig:reward2}\includegraphics[width=0.46\linewidth]{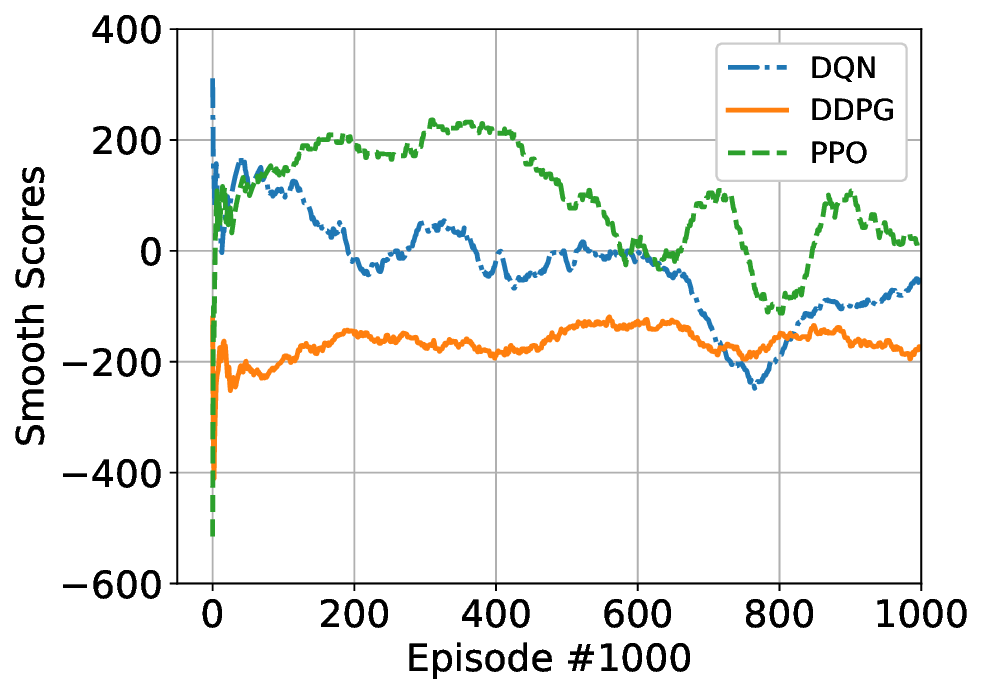}
    }
\caption{The training curves of the three agents: (a). the default reward function. (b). the reshaped reward function.}
\label{fig:reward_train}
\end{figure}

The training performance of the two reward functions and three RL agents is illustrated in Fig.~\ref{fig:reward_train}. In both figures, we notice that PPO reaches the highest training scores. It also achieves the largest performance increase from the default reward function to the reshaped reward function, among the three agents. This reveals that PPO is very sensitive to exploration encouragement and risk penalties, probably because of the usage of policy gradients. The DQN agent has lower training curves than the PPO. Its greedy search strategy makes it short-sighted, lacking the capability of exploration. DDPG has the lowest training rewards but with the most steady changes. This is because the actor-critic structure ensures steady learning progress which, however, costs time to achieve high training scores. The training results indicate that PPO is likely to be the best RL method for this autonomous driving study. However, a conclusion about the efficacy of reward shaping by comparing the two figures can not be drawn, since the two figures are defined by two different rewards. Further test study is needed to address the efficacy of reward shaping.

\subsection{Test Results}
In the test study, all trained agents were experimented on the same track separately and iterated for $1000$ episodes. For each test episode, the initial position of the car is also randomly sampled. There is no limitation on the number of steps. Instead, a test episode ends when the cumulative reward reaches the limits $R_{\mathrm{up}}=3000$ and $R_{\mathrm{down}}=-400$ which are the same as the training process. For each test episode, we use $T_{\mathrm{srv}}$ and $R_{\mathrm{cum}}$ to represent the total time of the episode and the corresponding cumulative reward, respectively. Then, we use the following three scores to evaluate the test performance of the three agents trained using the default reward and the reshaped reward functions.
\begin{itemize}
\item The out-of-track (o$^2$t) counts $N_{o^2t}$: the number of instants when the car drives out of the track;
\item The out-of-edge (oe) counts $N_{oe}$: the number of instants when the car drives close to the edge;
\item The surviving time $T_{\mathrm{srv}}$ (s): the total surviving time of the car over the $1000$ episodes.
\end{itemize}
The test scores of the three agents are presented in Table~\ref{tab: survival_time}.

\linespread{1.2}
\begin{table}[!ht]
\centering
\caption{The test scores of the three agents.
}\label{tab: survival_time}
\begin{tabular}{c|c|c|c|c}
\hline
Scores                                           & \multicolumn{1}{l|}{Reward Functions} & \multicolumn{1}{l|}{DQN}  & \multicolumn{1}{l|}{DDPG} & \multicolumn{1}{l}{PPO}  \\ \hline
\multirow{2}{*}{$N_{o^2t}$}                      & default                              & 0                         & 771                       & 721                      \\
                                                 & reshape                              & 0                         & 769                       & 5                        \\ \hline
\multirow{2}{*}{$N_{oe}$}                       & default                              & 47                        & 3172                      & 5323                     \\
                                                 & reshape                              & 42                        & 3156                      & 3177                     \\ \hline
\multicolumn{1}{l|}{\multirow{2}{*}{$T_{srv}$}} & default                              & \multicolumn{1}{l|}{1628} & \multicolumn{1}{l|}{1278} & \multicolumn{1}{l}{1654} \\
\multicolumn{1}{l|}{}                            & reshape                              & \multicolumn{1}{l|}{1693} & \multicolumn{1}{l|}{1328} & \multicolumn{1}{l}{3706} \\ \hline
\end{tabular}
\end{table}


\linespread{1}

%

From Tab.~\ref{tab: survival_time}, we can see that the reshaped reward in general leads to a longer survival time $T_{\mathrm{srv}}$ and fewer o$^2$t and oe counts. A longer survival time and fewer o$^2$t counts represent the safer execution of the autonomous driving task, while fewer oe counts indicate the better handling of risky events. Therefore, the overall results reveal the advantage of the proposed reward-shaping method based on encouragement and penalization for all three agents. Also, we can notice that the DQN agent achieves the best overall performance among all three agents. Meanwhile, PPO achieves the largest performance increase after reward shaping, which is similar to the training result discussed in Sec~\ref{sec:train_result}. The DDPG agent achieves mediocre performance which does not change much after reward shaping, which is consistent with the training results.

\begin{figure}[htbp] 
\centering
\subfloat[]{\includegraphics[height=3.2cm]{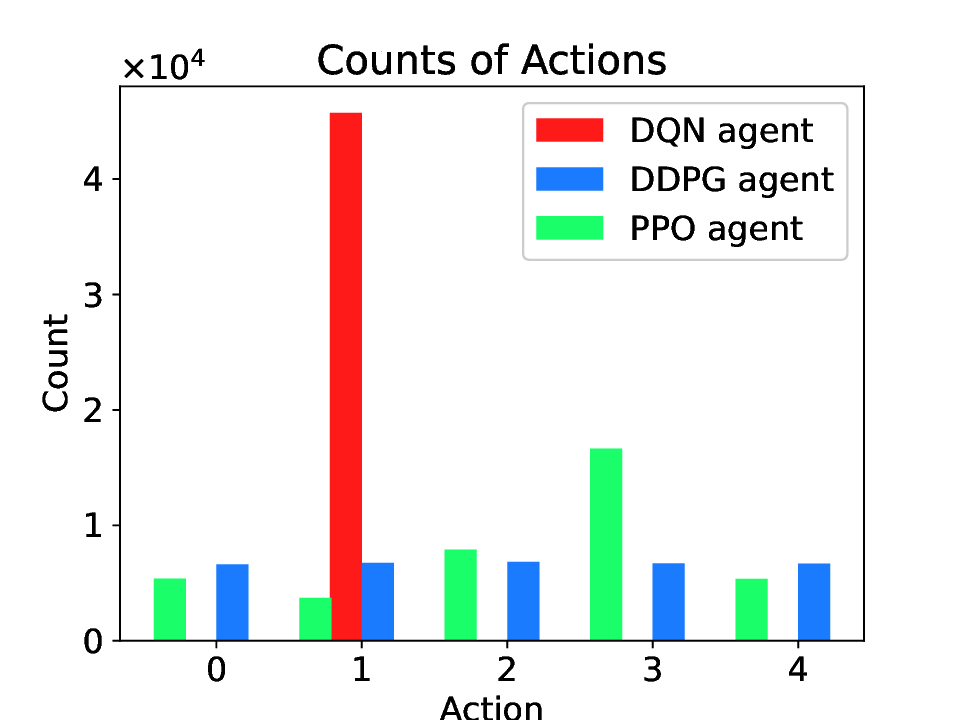}
    \label{fig:test:def:aver}}
\hfill
\subfloat[]{\includegraphics[height=3.2cm]{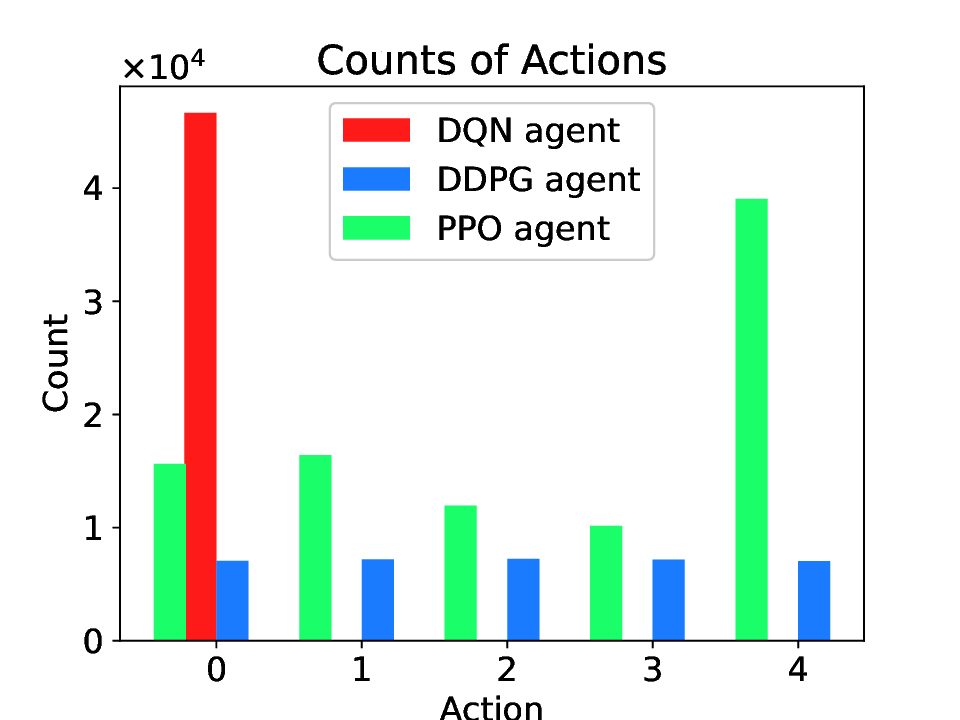}
    \label{fig:test:def:act}}
\caption{The counts of actions of the three RL agents where the number $0$ to $5$ denote the discrete actions \textit{no action}, \textit{steer left}, \textit{steer right}, \textit{accelerate}, and \textit{brake}, and (a) is for the default reward function and (b) is for the reshaped reward function.}
\label{fig:Test_default}
\end{figure}

Nevertheless, only the test scores are not sufficient to address the quality of the trained policies of the agents. We further investigate the action counts of the agents in the test to find out whether the agents are taking effective actions instead of lazy actions like \textit{no action}. Fig.~\ref{fig:Test_default} shows the total number of different actions that the three agents have taken during the test study. It can be noticed that the reshaped reward makes the DQN agent more conservative since it simply always takes \textit{no action} for all test steps. Therefore, the policy of the DQN agent is invalid even though it achieves the best scores in Tab.~\ref{tab: survival_time}. This implies a negative effect of the reshaped reward on DQN. On the contrary, PPO shows a more practical driving strategy and becomes more aggressive after reward shaping, since the number of \textit{braking} increases a lot. This indicates that it is taking practical actions to accomplish the driving task. More \textit{braking} actions mean that the PPO agent tries to prevent the car from going off the track or colliding with obstacles immediately when the agent accelerates. Different from DQN and PPO, a DDPG agent corresponds to a more even distribution of different actions, and this distribution does not change much after the reward shaping. To sum up, we can conclude that the encouragement of exploration and the penalization of risky driving positively contribute to the performance promotion of PPO, but negatively affect DQN. Meanwhile, their influence on DDPG is very limited.

\subsection{Discussion of Reward Shaping for Three RL Agents}

In this subsection, we summarize the results we have seen in the experimental studies and discuss how the proposed reward-shaping method can be extended for general autonomous driving applications. The results indicate that the DQN algorithm exhibits the lowest performance compared to the other algorithms. This is mainly because its greedy-search-based policy is \textit{short-sighted}, which makes it unsuitable to incorporate exploration- or risk-related rewards. As a result, the trained DQN agent cannot establish a steady policy for navigating the racing track and is prone to remain in the same position while taking no valid actions. 
Conversely, the DDPG algorithm, which employs continuous action spaces, can effectively learn to navigate the track by accruing positive rewards through successful road traversal. However, DDPG may require an extended number of episodes to achieve a decent performance due to its actor-critic structure. The effects of the exploration encouragement and the risk penalization are not obvious if the training is not sufficiently long. By contrast, the PPO algorithm exhibits excellent performance and great score promotion after the proposed reward shaping. This is mainly because the proximity guarantee of the policy updates ensures that the agent has steady policy promotions. Therefore, PPO is the most promising RL method that works with risk-aware reward shaping. Nevertheless, a drawback of PPO is that the agent halts the forward decisions too frequently, resulting in the agent moving forward at a lower speed than the DDPG agent. This is mainly due to the aggressiveness of the learning strategy of PPO. Additional reward terms can be introduced to penalize aggressive actions to make the agent more conservative. 

\section{Conclusion}\label{sec:con}

This study investigates how to use risk-aware reward shaping to improve the performance of three widely-used RL agents, namely DQN, DDPG, and PPO, in solving the autonomous driving problem in a simulated environment. We propose a risk-aware reward-reshaping function to encourage exploration and discourage driving risks. PPO implies the biggest promise to work with the proposed risk-aware rewards. Nevertheless, risk-aware reward shaping still requires heuristic knowledge to penalize the aggressive driving strategies of the PPO agents, which is a challenging topic. We will also incorporate additional rewards in complex forms, such as temporal logic specifications, to solve more complicated autonomous driving tasks.

\balance

\bibliographystyle{IEEEtran}
\bibliography{reference.bib}


\end{document}